\documentclass[letterpaper, 10pt, conference]{IEEEconf}
\IEEEoverridecommandlockouts
\usepackage{decar-common}
\usepackage{decar-dynamics}
\usepackage{decar-lie}
\usepackage{decar-post}

\newsavebox{\subfigbox}

\makeatletter
\AtBeginDocument{
  \check@mathfonts
}
\makeatother

\newcommand{\pynav}{\texttt{navlie}\xspace}

\title{\LARGE \bf \texttt{navlie}: A Python Package for State Estimation on Lie Groups}

\author{Charles Champagne Cossette, Mitchell Cohen, Vassili Korotkine, Arturo del Castillo Bernal,\\ Mohammed Ayman Shalaby, James Richard Forbes${^1}$%
\thanks{*This work was supported by the NSERC Discovery and Alliance programs (with Denso and ARA Robotics), as well as by the Innovation for Defence Excellence and Security (IDEaS) program from the Department of National Defence (DND). Any opinions and conclusions in this work are strictly those of the authors and do not reflect the views, positions, or policies of - and are not endorsed by - IDEaS, DND, or the Government of Canada.}%
\thanks{$^{1}$The authors are with the Department of
Mechanical Engineering, McGill University, 817 Sherbrooke St. W., Montreal, QC
H3A 0C3, Canada. Corresponding author is {\tt\small charles.cossette@mail.mcgill.ca}}%
}

\begin{document}
%
%
%
%
%
%
%
\def \myJournal {IEEE/RSJ International Conference on Intelligent Robots and Systems}
\def \myDoi {10.1109/LRA.2023.3268043}
\def \myPaperSiteName {IEEE Xplore}
\def \myPaperSiteLink {https://ieeexplore.ieee.org/document/10103620}
\def \myYear {2023}

\def \myPaperCitation{C. C. Cossette, M. Cohen, V. Korotkine, A. del Castillo Bernal,
Mohammed Ayman Shalaby, and J. R. Forbes ``Know What You Don't Know: Consistency in Sliding Window Filtering With Unobservable States Applied to Visual-Inertial SLAM,'' in \textit{IEEE Robotics and Automation Letters}, vol. 8, no. 6, pp. 3382-3389, June 2023.}


\begin{figure*}[t]

\thispagestyle{empty}
\begin{center}
\begin{minipage}{6in}
\centering
This paper has been accepted for publication at the \emph{\myJournal}. 
\vspace{1em}

This is the author's version of an article that has, or will be, published in this journal or conference. Changes were, or will be, made to this version by the publisher prior to publication.
\vspace{2em}


\vspace{2em}


\vspace{15cm}
\copyright \myYear \hspace{4pt}IEEE. Personal use of this material is permitted. Permission from IEEE must be obtained for all other uses, in any current or future media, including reprinting/republishing this material for advertising or promotional purposes, creating new collective works, for resale or redistribution to servers or lists, or reuse of any copyrighted component of this work in other works.

\end{minipage}
\end{center}
\end{figure*}
\newpage
\clearpage
\pagenumbering{arabic} 
\maketitle
\thispagestyle{empty}
\pagestyle{empty}

\fontdimen16\textfont2=\fontdimen17\textfont2
\fontdimen13\textfont2=5pt

\begin{abstract}The ability to rapidly test a variety of algorithms for an arbitrary state estimation task is valuable in the prototyping phase of navigation systems. Lie group theory is now mainstream in the robotics community, and hence estimation prototyping tools should allow state definitions that belong to manifolds. A new package, called \pynav, provides a framework that allows a user to model a large class of problems by implementing a set of classes complying with a generic interface. Once accomplished, \pynav provides a variety of on-manifold estimation algorithms that can run directly on these classes. The package also provides a built-in library of common models, as well as many useful utilities. The open-source project can be found at
\begin{center}
\url{https://github.com/decargroup/navlie}
\end{center}
\end{abstract}


\vspace{-2mm}
\section{Introduction}
\label{sec:intro}
To achieve full autonomy, robotic systems require the robot to maintain a belief of the current system state, given noisy sensor observations. Typically, this problem is solved using recursive Bayesian filters \cite{sarkka2013bayesian}, such as Kalman-like filters, or optimization-based smoothing methods. Variants of the Kalman filter often used for nonlinear state estimation include the Extended Kalman Filter (EKF) \cite[Ch. 5.2]{sarkka2013bayesian} and the sigma-point Kalman filter (SPKF) \cite[Ch. 5.6]{sarkka2013bayesian}, while smoothing methods rely on batch estimation \cite{Barfoot2022}. These traditional approaches assume that the system state is an element of Euclidean space. Difficulty arises in robotics applications, where the system state often evolves on a manifold.
In recent years, significant progress had been made to formulate robotic state estimation problems using tools from Lie theory \cite{Sola2018a}. Early applications of Lie groups to navigation typically focused on orientation estimation \cite{mahony2008nonlinear}, where attitude is represented as an element of the Special Orthogonal Group $SO(3)$. More recently, utilizing Lie groups has shown success in several robotics applications, including characterizing the uncertainty of poses \cite{Barfoot2014, mangelson2020characterizing}, addressing consistency issues in simultaneous localization and mapping (SLAM) problems \cite{heo2018consistent, brossard2018exploiting}, and inertial navigation problems \cite{brossard2021associating, brossard2020ai}. The use of Lie groups is also a crucial part of the invariant EKF \cite{barrau2016invariant}, and the equivariant filter \cite{van2020equivariant}.

Given these recent successes, several software packages have been developed to aid in designing on-manifold state estimators. The library \texttt{kalmanif}, based on \cite{Sola2018a}, provides a collection of Kalman filters on Lie groups. The OpenVINS codebase \cite{geneva2020openvins} provides a foundation for developing on-manifold visual-inertial estimators. Additionanlly, libraries developed for optimization-based state estimation such as Ceres \cite{Agarwal_Ceres_Solver_2022}, GTSAM \cite{dellaert2012factor}, and g2o \cite{grisetti2011g2o} also support on-manifold estimation. These libraries are all implemented in C++, allowing for real-time application, at the cost of making their use for rapid algorithm prototyping more difficult. The library UKF-M \cite{bonnabelukf} provides Python and \MATLAB implementations of an on-manifold unscented Kalman filter (UKF). Recently, the PyTorch-based package \texttt{PyPose} \cite{wang2022pypose} was released, providing a collection of differentiable estimation and control tools, meant to connect learning-based models with physics-based optimization.

The main contribution of this paper is to present the library \pynav, which contains a collection of on-manifold estimation algorithms and tools for rapid prototyping. At the time of writing, \pynav  has the following features:
\begin{itemize}
    \item Generic implementations of common filtering algorithms, namely the EKF, iterated EKF, invariant EKF, multiple sigma-point Kalman filters, including the unscented, spherical cubature, and Gauss-Hermite filters, and the interacting multiple model filter.
    \item Batch estimation in a maximum a posterori (MAP) framework.
    \item Implementations of common Lie groups such as $SO(2)$, $SO(3)$, $SE(2)$, $SE(3)$, $SE_2(3)$ and $SL(3)$.
    \item A collection of common process and measurement models.
    \item A preintegration module for linear, wheel odometry, and IMU process models.
    \item Utilities for numerical differentiation, plotting, error and consistency evaluation, and conducting Monte-Carlo experiments.  
\end{itemize}
This extended feature set, accessible with the convenience of Python, makes \pynav   unique.



\vspace{-1mm}
\section{Preliminaries}
\vspace{-1mm}
\label{sec:preliminaries}
\subsection{Lie groups}

A Lie group $G$ is a smooth manifold whose elements, given a group operation $\circ: G \times G \to G$, satisfy the group axioms \cite{Sola2018a}. The application of this operation to two arbitrary Lie group elements $\mc{X, Y} \in {G}$ is written as $\mc{X} \circ \mc{Y} \in G$. For any $G$, there exists an associated Lie algebra $\mathfrak{g}$, a vector space identifiable with elements of $\mathbb{R}^m$, where $m$ is referred to as the degrees of freedom of $G$. The Lie algebra is related to the group through the exponential and logarithmic maps, denoted $\exp: \mathfrak{g} \to G$ and $\log: G \to \mathfrak{g}$. The ``vee'' and ``wedge'' operators are denoted $(\cdot)^\vee: \mathfrak{g} \to \mathbb{R}^m$ and $(\cdot)^\wedge: \mathbb{R}^m \to \mathfrak{g}$, and are used to associate group elements with vectors with 
\beq \nonumber
\mc{X} = \exp(\mbs{\xi}^\wedge) \triangleq \Exp(\mbs{\xi}), \qquad \mbs{\xi} = \log(\mc{X})^\vee \triangleq   \Log(\mc{X}),
\eeq 
where $\mc{X} \in G, \mbs{\xi} \in \mathbb{R}^m$, and the shorthand notation $\Exp: \mathbb{R}^m \to G$ and $\Log:G \to \mathbb{R}^m$ has been defined. The most common Lie groups appearing in robotics are $SO(n)$, representing rotations in $n$-dimensional space, $SE(n)$, representing poses, and $SE_2(3)$ representing ``extended'' poses that also contain velocity information. In these cases, the element $\mc{X}$ is an invertible matrix and the group operation $\circ$ is regular matrix multiplication.

\subsubsection{$\oplus$ and $\ominus$ operators}
Estimation theory for vector-space states and Lie groups can be elegantly aggregated into a single mathematical treatment by defining generalized ``addition'' $\oplus: G \times \mathbb{R}^m \to G$ and ``subtraction'' $\ominus:G \times G \to \mathbb{R}^m$ operators, whose precise definitions are problem dependant. For example, possible implementations include 
\beq
\begin{array}{ll}
    \mc{X} \oplus \delta \mbf{x} = \mc{X} \circ \Exp(\delta \mbf{x}) & \text{(Lie group right)},\\
    \mc{X} \oplus \delta \mbf{x} =  \Exp(\delta \mbf{x}) \circ  \mc{X} & \text{(Lie group left)},\\
    \mbf{x} \oplus \delta \mbf{x} = \mbf{x} + \delta \mbf{x} & \text{(vector space)}, \label{eq:oplus}
\end{array}
\eeq
for addition and, correspondingly,
\beq
\begin{array}{ll}
    \mc{X} \ominus \mc{Y} = \Log(\mc{Y}^{-1} \circ \mc{X}) & \text{(Lie group right)},\\
    \mc{X} \ominus \mc{Y} =  \Log(\mc{X} \circ \mc{Y}^{-1})& \text{(Lie group left)},\\
    \mbf{x} \ominus \mbf{y} = \mbf{x} - \mbf{y} & \text{(vector space)}, \label{eq:ominus}
\end{array}
\eeq
for subtraction. This abstraction is natural since a linear vector space technically qualifies as a Lie group with regular addition $+$ as the group operation. 


\subsubsection{Derivatives on Lie groups}
Again following \cite{Sola2018a}, the Jacobian of a function $f: G \to G$, taken with respect to $\mc{X}$ can be defined as 
\beq \nonumber
\left.\frac{D f(\mc{X})}{D \mc{X}}\right|_{\bar{\mc{X}} }\triangleq \left.\frac{\p f(\bar{\mc{X}} \oplus \delta \mbf{x}) \ominus f(\bar{\mc{X}})}{\p \delta \mbf{x}}\right|_{\delta \mbf{x} = \mbf{0}},
\eeq
where it should be noted that the function  $f(\bar{\mc{X}} \oplus \delta \mbf{x}) \ominus f(\bar{\mc{X}})$ of $\delta \mbf{x}$ has $\mathbb{R}^m$ as both its domain and codomain, and can thus be differentiated using any standard technique. With the above general definition of a derivative, it is easy to define the so-called \emph{ Jacobian of }$G$ as $\mbf{J} = D \Exp(\mbf{x})/ D \mbf{x}$, where left/right group Jacobians are obtained with left/right definitions of $\oplus$ and $\ominus$. 
\begin{figure}[t]
\centering 
\includegraphics[width=0.97\linewidth,clip,trim={0.75cm 3cm 10cm 0}]{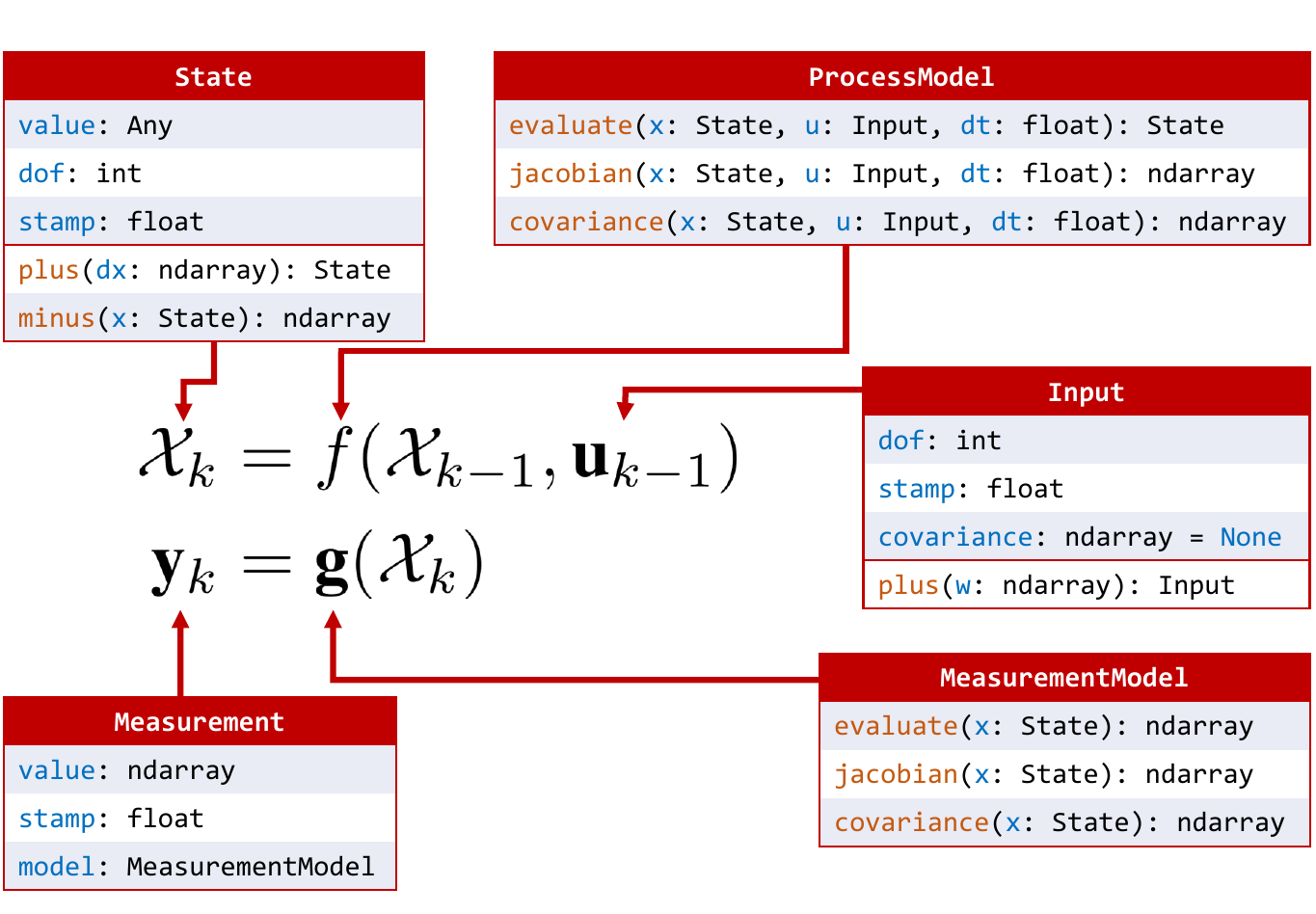}
\vspace{-3mm}
\caption{Class diagram for the abstract types used in \pynav. The methods shown should be implemented by the user.}
\label{fig:class_diagram}
\end{figure}
\vspace{-1mm}
\section{Estimation Algorithms on Manifolds}
\vspace{-2mm}
\label{sec:on_manifold_estimation}

\pynav operates on implementations of states, process, and measurement models given in the standard form of 
\begin{align}
    \mc{X}_{k} &= f(\mc{X}_{k-1}, \mbf{u}_{k-1}) \oplus \mbf{w}_{k-1}, \label{eq:proc_model}\\
    \mbf{y}_{k} &= \mbf{g}(\mc{X}_{k}) + \mbf{v}_{k}, \label{eq:meas_model}
\end{align}
where $\mbf{u}_{k-1}$ is a generic process model input, and $\mbf{w}_{k-1} \sim \mc{N}(\mbf{0}, \mbf{Q}_{k-1})$, $ \mbf{v}_k \sim \mc{N}(\mbf{0}, \mbf{R}_k)$ are ``additive'' Gaussian noises. Figure \ref{fig:class_diagram} shows the key classes and abstract methods that represent the system model. For the state, the user must implement $\oplus$ and $\ominus$ functions. For the process model, the user must implement the function $f$, and a computation of $\mbf{Q}_{k-1}$. Finally, for the measurement model, the user must implement $\mbf{g}$ and $\mbf{R}_k$. Jacobians can optionally be implemented by the user for maximum performance, but a default finite-difference method is used otherwise.

The remainder of this section the mathematical describes the details of the algorithms found in \pynav. The filter classes in this package are \emph{stateless}, which allows the user to switch and combine estimators from timestep to timestep, or to use specific estimators for specific measurement types.
\vspace{-1mm}
\subsection{Extended Kalman Filter}
\vspace{-1mm}
The EKF in \pynav follows a standard covariance-form implementation, and follows a predict-correct structure. The EKF is initialized with a mean $\mchat{X}_0$ and covariance $\mbfhat{P}_0$, and the prediction step is given by 
\begin{align*}
    \mbf{F}_{k-1} &= \left.\frac{D f(\mc{X}, \mbf{u}_{k-1})}{D \mc{X}}\right|_{\mchat{X}_{k-1}}, \\
    \mccheck{X}_k &= f(\mchat{X}_{k-1}, \mbf{u}_{k-1}), \\
    \mbfcheck{P}_k &= \mbf{F}_{k-1} \mbfhat{P}_{k-1} \mbf{F}_{k-1}^\trans + \mbf{Q}_{k-1},
\end{align*}
where $\check{\left(\cdot\right)}$ and $\hat{\left(\cdot \right)}$ refer to predicted and corrected variables, respectively.
The correction step is given by 
\begin{align*}
    \mbf{G}_k &= \left.\frac{D \mbf{g}(\mc{X})}{D \mc{X}}\right|_{\mccheck{X}_k}, \\
    \mbf{K} & \triangleq \mbfcheck{P}_k \mbf{G}_k (\mbf{G}_k \mbfcheck{P}_k \mbf{G}_k^\trans + \mbf{R}_k)^{-1}, \\
    \mbf{z} &\triangleq \mbf{y}_k - \mbf{g}(\mccheck{X}_k), \\
    \mchat{X}_k &= \mccheck{X}_k \oplus \mbf{K}_k\mbf{z} ,\\
    \mbfhat{P}_k &= (\mbf{1} - \mbf{K} \mbf{G}_k) \mbfcheck{P}_k.
\end{align*}
\subsection{Iterated Extended Kalman Filter}
The iterated EKF treats the correction step as a nonlinear least squares problem solved using Gauss-Newton in an iterative manner \cite{bourmaud2016intrinsic}. In \pynav, the iterated EKF inherits the EKF prediction step, but performs a correction step by first computing
\beq
\begin{aligned}
    \mbf{K} &\triangleq \mbf{J}_k \mbfcheck{P}_k \mbf{J}_k^\trans \mbf{S}_k^\trans ( \mbf{G}_k\mbf{J}_k \mbfcheck{P}_k \mbf{J}_k^\trans \mbf{G}_k^\trans+ \mbf{R}_k)^{-1},\\
    \mbf{z} &\triangleq \mbf{y}_k - \mbf{g}(\mchat{X}_k) + \mbf{G}_k \mbf{J}_k(\mchat{X}_k \ominus \mctilde{X}_k), \\
    \delta \mbf{x}_k & = \mbf{K} \mbf{z} - \mbf{J}_k(\hat{\mc{X}}_k \ominus \mctilde{X}_k) ,\label{eq:iter_ekf}
\end{aligned}
\eeq
and then updating the current estimate $\mchat{X}_k \gets \mchat{X}_k \oplus \delta \mbf{x}_k$, initialized with $\mchat{X}_k \gets \mccheck{X}_k$. Equations \eqref{eq:iter_ekf} are continuously iterated until convergence.
\vspace{-1mm}
\subsection{Invariant Extended Kalman Filter}
The invariant EKF \cite{barrau2016invariant} defines the state and innovation such that the process and measurement model Jacobians are state estimate independent for a specific form of process and measurement models. The invariant EKF exploits measurement models with one of two forms, 
\begin{align*}
    \mbf{y}_k &= \mc{X}_k \cdot \mbf{b}_k + \mbf{v}_k, \qquad \text{(left-invariant)}\\
    \mbf{y}_k &= \mc{X}_k^{-1} \cdot \mbf{b}_k + \mbf{v}_k  \qquad \text{(right-invariant)}
\end{align*}
where $\mbf{b}_k \in \mathbb{R}^n$ is an arbitrary known vector and $\cdot$ denotes the \emph{group action} \cite{Sola2018a}. When the measurement model is of the right-invariant form, the invariant EKF defines the innovation to be used in the filter as 
\begin{align*}
\mbf{z} &= \check{\mc{X}}\cdot(\mbf{y} - \mbf{g}(\check{\mc{X}}))\\
&= \check{\mc{X}}\cdot(\mbf{g}(\mc{X}) +
\mbf{v} - \mbf{g}(\check{\mc{X}}))\\
&\approx \check{\mc{X}}\cdot( \mbf{g}(\check{\mc{X}})
+ \mbf{G}\delta \mbf{x} + \mbf{v}
- \mbf{g}(\check{\mc{X}}))\\
&= \check{\mc{X}}\cdot\mbf{G}\delta \mbf{x}
+ \check{\mc{X}}\cdot\mbf{v}
\end{align*}
where a Taylor series expansion of $\mbf{g}(\mc{X})$ was used. The result is that the Jacobian of the innovation $\mbf{z}$ is given by $\check{\mc{X}}\cdot\mbf{G}$, which will be state-independent if a right-invariant measurement model is used with a left-definition of the $\oplus$ operator. Similarly, when the measurement model is of the left-invariant form, the innovation used in the invariant EKF is written
\begin{align*} 
\mbf{z} &= \mccheck{X}^{-1}\cdot(\mbf{y} - \mbf{g}(\mccheck{X})) \approx \check{\mc{X}}^{-1}\cdot\mbf{G}\delta \mbf{x}
+ \check{\mc{X}}^{-1}\cdot\mbf{v},
\end{align*}
and the Jacobian of the left-invariant innovation is thus $\check{\mc{X}}^{-1}\cdot\mbf{G}$, which will be state-independent with a left-invariant measurement model and right $\oplus$ definition. The implementation of the invariant EKF in \pynav allows users to reuse existing measurement models with the specific definitions of the innovation presented above. The construction of the left- or right- invariant innovations is done automatically when possible.
\vspace{-1mm}
\subsection{Sigma-Point Kalman Filter}
The sigma-point transform generates a set of sigma-points from a prior distribution, then
passes them through a non-linearity to estimate the mean and covariance of the transformed distribution \cite{sarkka2013bayesian}. In \pynav, sigma points can be generated by unscented (UKF), spherical cubature (CKF), and Gauss-Hermite (GHKF) transforms.

\subsubsection{Prediction Step}
For the sigma-point filters, noise is assumed to be additive to the input such that the process model is $f(\mc{X}_{k-1}, \mbf{u}_{k-1} + \mbf{w}_{k-1})$ with $\mbf{w}_k \sim \mc{N}(\mbf{0}, \mbf{Q}^u_k)$. The sigma points are then calculated by 
\begin{align*}
    {\mbftilde{P}}_{k-1}
    &=  \diag(\hat{\mbf{P}}_{k-1}, \mbf{Q}^u_{k-1}) \triangleq \mbf{L}\mbf{L}^\trans,\\
    \Big[{\delta \mbf{x}^{(i)}}^\trans \;\; \delta {\mbf{w}^{(i)}}^\trans\Big]^\trans &= \mbf{L}\mbs{\xi}^{(i)} ,
\end{align*}
where $\mbf{L}$ is a Cholesky decomposition,  $\mbs{\xi}^{(i)}$ is the $i^{\textrm{th}}$ unit sigma point generated by one of the transforms, and $w^{(i)}$ is a corresponding weight. Sigma points are propagated with
\bdis
    \check{\mc{X}}_{k}^{(i)} = f(\hat{\mc{X}}_{k-1}\oplus \delta \mbf{x}^{(i)}, {\mbf{u}}_{k-1} + \delta \mbf{w}^{(i)}) . 
\edis
The mean $\check{\mc{X}}_{k}$ is computed in an iterated manner, where the first of the propagated states is set as the initial mean estimate, $\mccheck{X}_k \gets \mccheck{X}_k^{(1)}$, and the error with the rest of the propagated states is computed with a weighted mean. The mean is updated until a convergence criterion is met, 
\bdis
    \delta \mbf{x}_j = \sum_{i} w^{(i)} \check{\mc{X}}_{k}^{(i)}\ominus\check{\mc{X}}_{k}, \quad
    \check{\mc{X}}_{k} \gets \check{\mc{X}}_{k} \oplus\delta \mbf{x}_j .
\edis
After convergence, the covariance is computed with
\begin{align*}
    \mbfcheck{P}_k & = \sum_{i} w^{(i)} \left(\check{\mc{X}}_{k}^{(i)}\ominus\check{\mc{X}}_{k}\right)\left(\check{\mc{X}}_{k}^{(i)}\ominus\check{\mc{X}}_{k}\right)^\trans.
\end{align*}
\subsubsection{Correction Step}

The correction step is handled the same way as in \cite[Sec. II -- C]{bonnabelukf}. At the time of writing, sigma points are generated for the state only since the noise is assumed to be additive to the output in the measurement model. Thus,
\begin{align*}
    \mbf{y}_{k}^{(i)} &= \mbf{g}(\check{\mc{X}}_{k}\oplus \delta \mbf{x}^{(i)}), \quad \mbfbar{y}_k = \sum_i w^{(i)} \mbf{y}_{k}^{(i)}, \\
    \mbf{P}_{yy} &= \sum_i w^{(i)} \big(\mbf{y}_{k}^{(i)}-\mbfbar{y}_k \big)\big(\mbf{y}_{k}^{(i)}-\mbfbar{y}_k \big)^\trans +\mbf{R}_k, \\
    \mbf{P}_{xy} &= \sum_i w^{(i)} \delta\mbf{x}^{(i)}\big(\mbf{y}_{k}^{(i)}-\mbfbar{y}_k \big)^\trans, \\
    \mbf{K} &= \mbf{P}_{xy}\mbf{P}_{yy}\inv ,\\
    \hat{\mc{X}}_{k} &=  \hat{\mc{X}}_{k} \oplus \mbf{K} \left(\mbf{y}_k - \mbfbar{y}_k \right), \quad \mbfhat{P}_k = \mbfcheck{P}_k -  \mbf{K}\mbf{P}_{yy} \mbf{K}^\trans.
\end{align*}

\subsection{Batch Estimation via Nonlinear Least Squares}
Batch estimation estimates the history of states from time $t= t_0$ to $t=t_K$, written as $\mc{X} = \mc{X}_{0:K} = \left\{\mc{X}_0, \ldots, \mc{X}_K \right\}$, that maximizes the posterior probability density function given a prior $\tilde{\mc{X}}_0$ and all the inputs and 
measurements received. Finding the MAP estimate can be equivalently posed as a weighted nonlinear least-squares problem as \cite{Barfoot2014}
\begin{multline*} 
    \hat{\mc{X}} = \arg \min_{\mc{X}} \frac{1}{2} \norm{\mc{X}_0 \ominus \check{\mc{X}}_0}^2_{\mbfcheck{P}} \\+ \frac{1}{2}\sum_{k=1}^{K} \left(\norm{\mc{X}_k \ominus f(\mc{X}_{k-1}, \mbf{u}_{k-1})}^2_{\mbf{Q}_{k-1}} + \norm{\mbf{y}_k - \mbf{g}(\mc{X}_k)}^2_{\mbf{R}_k} \right),
\end{multline*}
which is a combination of a prior, process, and measurement errors. \pynav provides on-manifold implementations of Gauss-Newton and Levenberg Marquardt \cite{Madsen2004} to solve nonlinear least squares problems that operate on generic error terms. Additionally, \pynav implements standard forms of prior, process, and measurement errors, allowing users to easily construct and solve batch problems utilizing any chosen state, process, and measurement definitions. 

%


\subsection{Interacting Multiple-Model Filter}
The Interacting Multiple Model Filter (IMM) 
\cite{blom1988interacting}
is a method for handling systems with discrete modes, modelled as a finite-state Markov chain with some probability transition matrix. 
The discrete modes can, for instance, correspond to different hypotheses for the process model. In this case, the overarching process model may be written
\bdis
    \mc{X}_k = f(\mc{X}_{k-1}, \mbf{u}_{k-1}, \theta_k),
\edis
where $\theta_k \in \{1, \dots, N\}$ is modelled as a finite-state Markov chain. The values of $\theta_k$ correspond to a different process model, such that $\mc{X}_k = f_{\theta_k}(\mc{X}_{k-1}, \mbf{u}_{k-1})$. 

A key step in the IMM algorithm is the computation of means and covariances of Gaussian mixtures. For Gaussian mixtures defined on a vector space, the means and covariances are straightforward to compute. However, for a Lie group, the procedure to compute Gaussian mixtures involves expressing all of the components in the same tangent space and is more involved.
\pynav seamlessly handles the manifold structure using the approach in \cite{cesip2017mixture}, reparametrizing the Gaussian mixture components about the component with the highest weight, carrying out the mixture in the corresponding tangent space, and projecting back to the manifold.

\section{Other on-manifold utilities} 
\label{sec:other_utilities}
\subsection{Finite-difference and complex-step differentiation}
\pynav also includes general-purpose numerical differentiation tools based on both finite differencing and the complex step \cite{cossette2020complex}. The $i^\mathrm{th}$ column of the Jacobian of $f(\mc{X})$ can be approximated with a forward difference procedure
\beq \nonumber
\left.\frac{D f(\mc{X})}{D \mc{X}}\right|_{\bar{\mc{X}} }\mbf{e}_i\approx \frac{f(\bar{\mc{X}} \oplus h \mbf{e}_i) \ominus f(\bar{\mc{X}})}{h},
\eeq
or using a complex step
\beq \nonumber
\left.\frac{D f(\mc{X})}{D \mc{X}}\right|_{\bar{\mc{X}} }\mbf{e}_i\approx \frac{\mathrm{Imag}\left(f(\bar{\mc{X}} \oplus h j \mbf{e}_i) \ominus f(\bar{\mc{X}})\right)}{h}, 
\eeq
where $j=\sqrt{-1}$ and $\mbf{e}_i \in \mathbb{R}^m$ is all zeros except for a 1 in the $i^\mathrm{th}$ element.
\subsection{Interpolation}
\pynav also provides a general interpolation method that works on any state implementation with well-defined $\oplus$ and $\ominus$ operators. Given two states $\mc{X}_k = \mc{X}(t_k),\; \mc{X}_{k-1} = \mc{X}(t_{k-1})$, an interpolated state at time $t$ can be obtained from 
\beq
\begin{aligned}
    \alpha &= (t - t_{k-1})/(t_k - t_{k-1}) \in [0,1], \\
    \delta \mbf{x} &= \mc{X}_k \ominus \mc{X}_{k-1}, \hspace{3mm} \mc{X}(t) = \mc{X}_{k-1} \oplus \alpha \delta \mbf{x}.
\end{aligned} \label{eq:interp}
\eeq
As usual, \pynav provides a single implementation of the above that works with any state implementation complying with the abstract \texttt{State} interface.
\begin{figure}
    \centering
    \includegraphics[width=0.9\linewidth, clip, trim={0 0.7cm 0 2.05cm}]{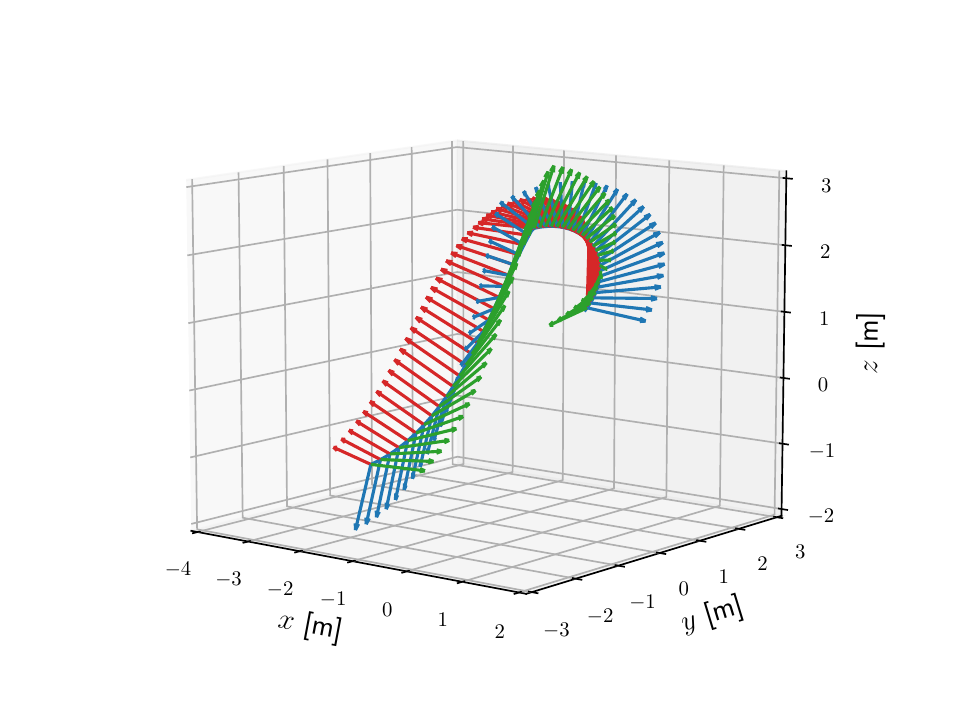}
    \vspace{-2mm}
    \caption{$SE(3)$ pose interpolation using \eqref{eq:interp}.}
    \label{fig:pose_interp}
    \vspace{-0.1cm}
\end{figure}
\subsection{Composite States} A \emph{composite group} or \emph{bundle} refers to a collection of states serving as a single concatenated object of the form
\beq \nonumber
\mbc{X} = (\mc{X}_1, \ldots \mc{X}_N) \in G_1 \times \ldots \times G_N.
\eeq
Defining $\delta \mbf{x} = [\delta \mbf{x}_1^\trans \; \ldots \; \delta \mbf{x}_N^\trans]^\trans$, the  $\oplus$ operator for this new state is defined elementwise by
\beq \nonumber
\mbc{X} \oplus \delta \mbf{x} = (\mc{X}_1 \oplus \delta \mbf{x}_1, \; \ldots \;, \mc{X}_N \oplus \delta \mbf{x}_N),
\eeq
and a similar definition applies to $\ominus$. The \texttt{CompositeState} class in \pynav allows users to freely combine and nest state implementations into a hierarchy of more complicated states, that all still comply with the \texttt{State} interface.

\vspace{-2mm}
\subsection{Preintegration}
\label{sec:preintegration}
\pynav provides preintegration classes for several common process models in robotics. For example, \pynav provides the ability to preintegrate any linear process model
\beq \nonumber
\mbf{x}_k = \mbf{F}_{k-1} \mbf{x}_{k-1} + \mbf{L}_{k-1}\mbf{u}_{k-1}.
\eeq
This is done by direct iteration to give
\begin{align*}
\mbf{x}_j &= \left(\prod_{k=i}^{j-1} \mbf{F}_k\right) \mbf{x}_i + \sum_{k=i}^{j-1} \left(\prod_{\ell = k + 1}^{j-1} \mbf{F}_\ell\right) \mbf{L}_k \mbf{u}_k \\ 
&\triangleq \mbf{F}_{ij} \mbf{x}_i + \Delta \mbf{x}_{ij}. \label{eq:linear_preint}
\end{align*}
Body-frame velocity-input process models of the form 
\beq \nonumber
\mbf{X}_k = \mbf{X}_{k-1} \Exp(\Delta t \mbf{u}_{k-1}),
\eeq 
where $\mbf{X}_k$ belongs to a Lie group such as $SO(n)$ or $SE(n)$, can also be preintegrated with \pynav, again done by direct iteration to produce 
\beq \nonumber
\mbf{X}_j = \mbf{X}_i \underbrace{\prod_{k=i}^{j-1} \Exp(\Delta t\mbf{u}_k)}_{\triangleq \Delta \mbf{X}_{ij}}.
\eeq
Finally, IMU preintegration is also included and is the first open-source Python implementation to do so directly on the $SE_2(3)$ manifold. Following \cite{Barfoot2022, brossard2021associating}, discrete-time IMU kinematics are of the form 
\beq \nonumber
\mbf{T}_{k} = \mbf{G}_{k-1} \mbf{T}_{k-1} \mbf{U}_{k-1},
\eeq 
where $\mbf{T}_k \in SE_2(3)$ and $\mbf{G}_k, \mbf{U}_k$ are $5\times5$ matrices constructed from gravity terms and IMU measurements. This is easily preintegrated with 
\begin{align*}
\mbf{T}_{{j}} &= \left(\prod_{k=i}^{j-1} \mbf{G}_{k-1}\right) \mbf{T}_{i} \left(\prod_{k=i}^{j-1} \mbf{U}_{k-1}\right) \triangleq \Delta \mbf{G}_{ij} \mbf{T}_i \Delta \mbf{U}_{ij}
\end{align*}
In all cases, \pynav handles noise propagation through the preintegration process, and also provides the ability to augment these models with input bias estimates.

\section{Example Usage}
\label{sec:examples}
\begin{figure}
    \centering
    \includegraphics[width=0.98\linewidth]{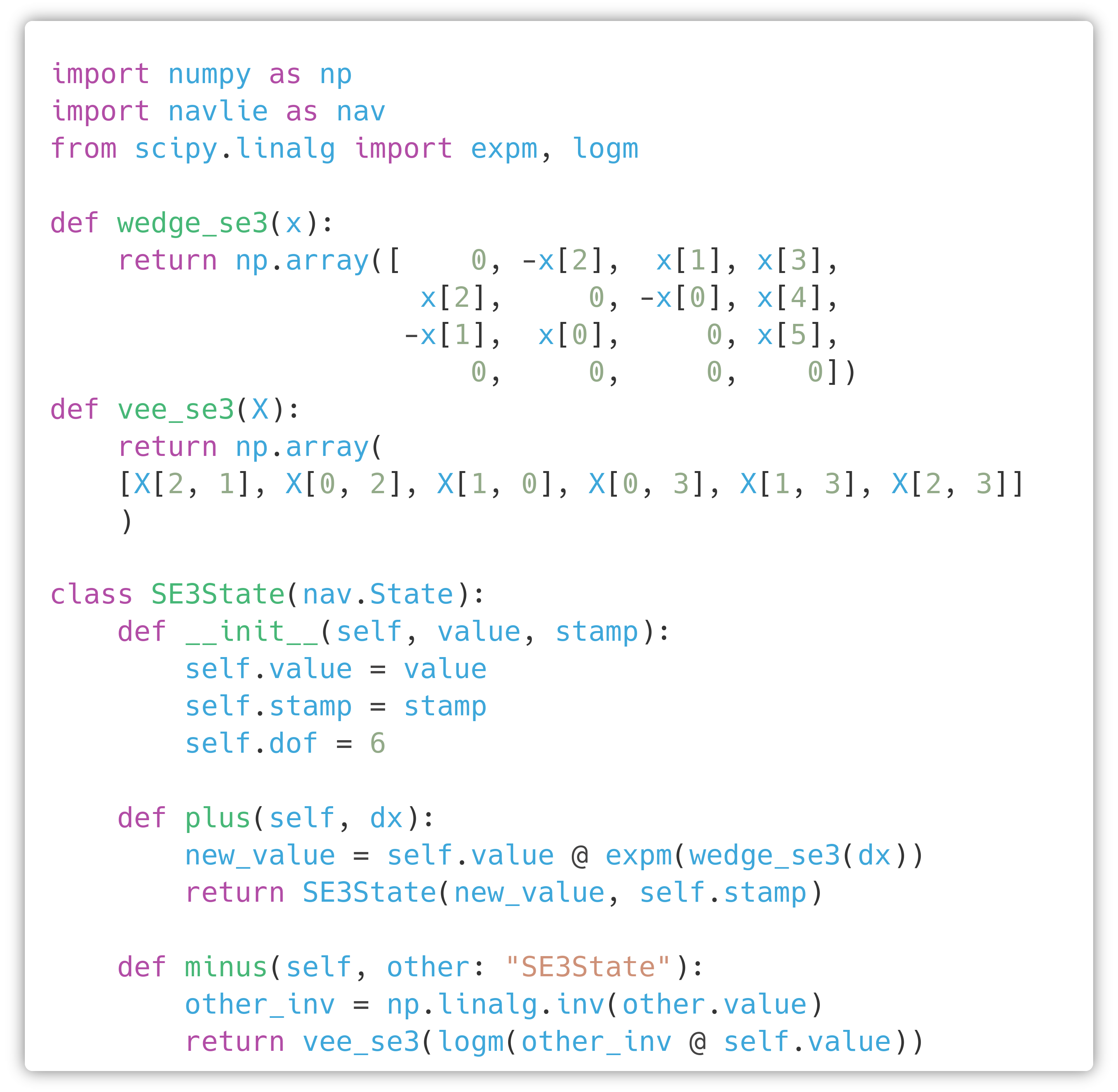}
    \vspace{-4mm}
    \caption{A minimal implementation of a state belonging to the $SE(3)$ manifold.}
    \label{fig:state_snippet}
    \vspace{-3mm}
\end{figure}

This section will show, concretely, an example implementation of a problem in \pynav. Consider the problem of estimating the pose represented as an element of $SE(3)$ given by
    \beq \nonumber
        \mbf{T} = 
            \begin{bmatrix}
                \mbf{C}_{wb} & \mbf{r}_w\\
                \mbf{0} & 1
            \end{bmatrix} \in SE(3),
    \eeq
where $\mbf{C}_{wb}$ is the direction-cosine-matrix (DCM) describing the attitude of a robot's body frame $\mc{F}_b$ relative to a world frame $\mc{F}_w$, and  $\mbf{r}_w$ is the position of the robot resolved in the world frame. In the considered example, a right-perturbation is chosen, and the definitions of the $\oplus$ and $\ominus$ operators are given by the ``Lie group right'' entries in \eqref{eq:oplus} and \eqref{eq:ominus} with  $\delta \mbf{x} \in \mathbb{R}^6$. Figure~\ref{fig:state_snippet} shows how this state definition can be constructed using the \texttt{State} interface in \pynav.

\begin{figure}[t]
    \centering
    \includegraphics[width=0.98\linewidth]{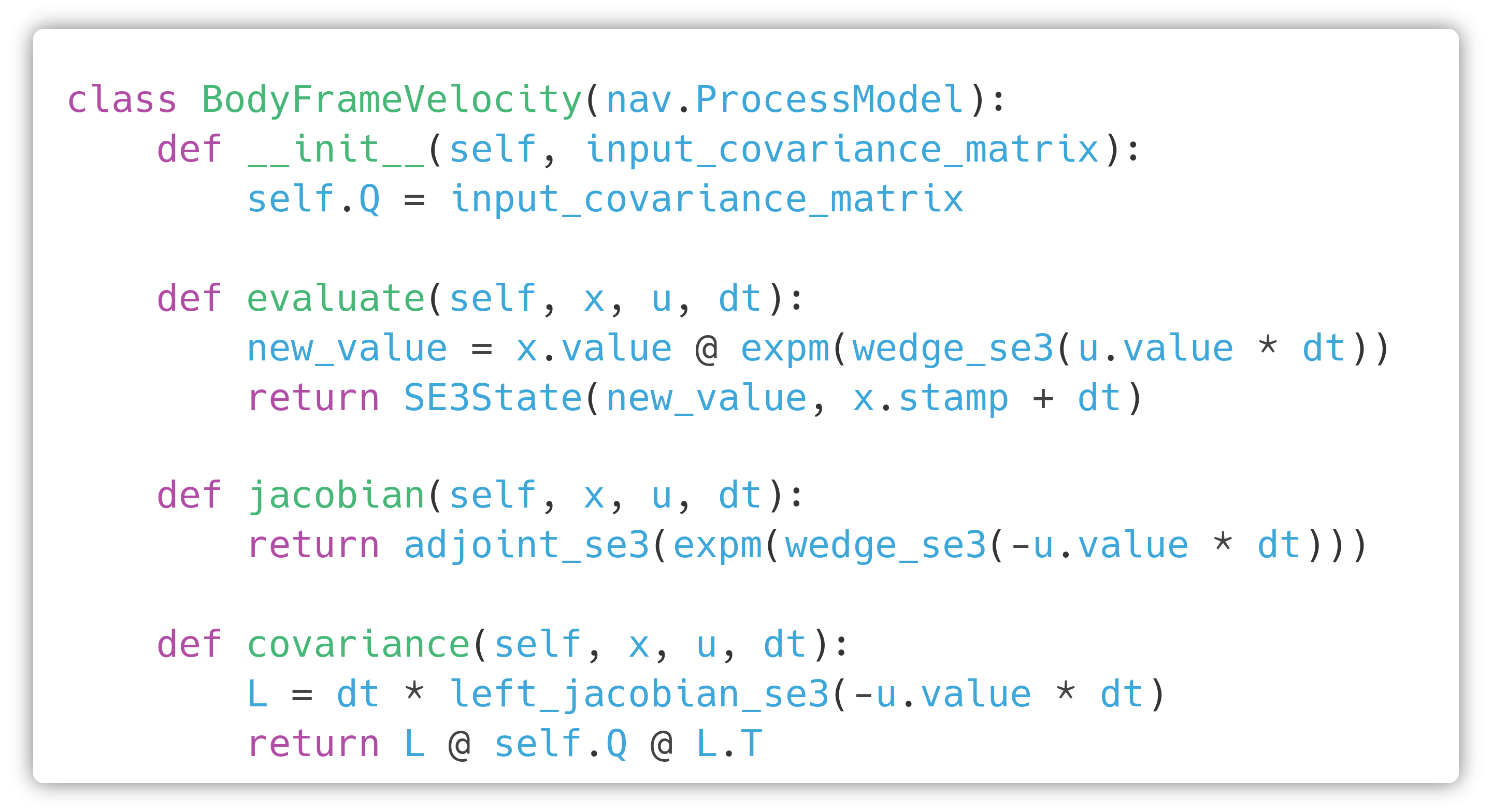}
    \vspace{-4mm}
    \caption{Implementation of the body-frame-velocity input process model in \eqref{eq:body_frame_velocity} using the \pynav framework. Standard implementations of the adjoint matrix operator and the left Jacobian for $SE(3)$ are omitted.}
    \label{fig:process_snippet}
    \vspace{-3mm}
\end{figure}

\begin{figure}[t]
    \centering
    \includegraphics[width=0.98\linewidth]{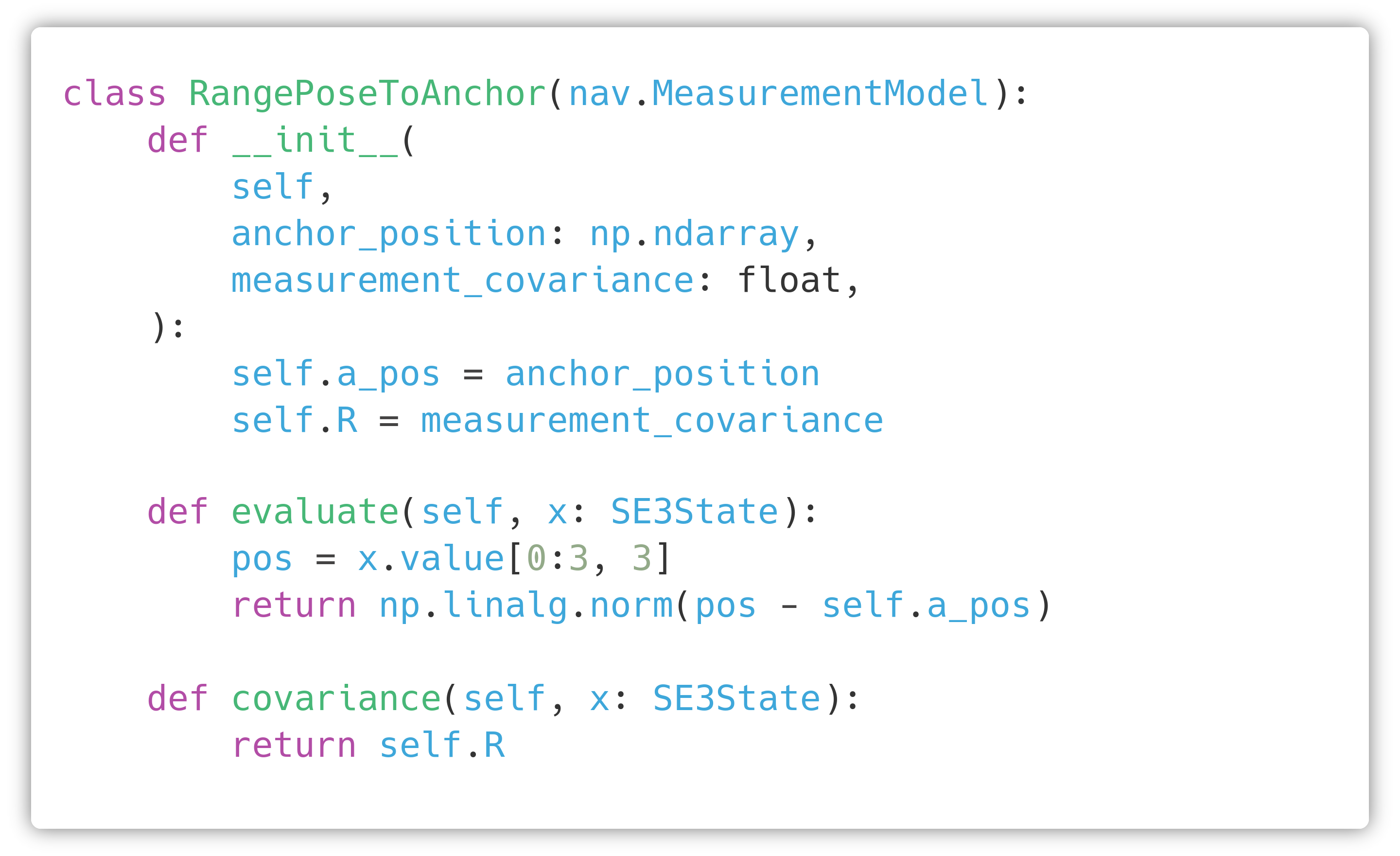}
    \vspace{-4mm}
    \caption{Implementation of a ranging measurement model to a single landmark as described by \eqref{eq:range_pose_to_anchor}. Here, the user has not implemented the \texttt{jacobian} method, meaning \pynav automatically uses finite difference.}
    \label{fig:meas_snippet}
\end{figure}

The robot has access to measurements of $\mbf{u}_k = \begin{bmatrix}\mbs{\omega}_{b_k}^\trans  &\mbf{v}_{b_k}^\trans  \end{bmatrix}^\trans$ 
where the translational velocity $\mbf{v}_b$ and angular velocity $\mbs{\omega}_b$ of the robot are both resolved in the body frame. The corresponding process model is
    \begin{align}
        \mbf{T}_k & = \mbf{T}_{k-1} \Exp \left(\Delta t \left(\mbf{u}_{k-1} + \mbf{w}_{k-1} \right) \right),\label{eq:body_frame_velocity}
    \end{align}
where $\mbf{w}_k \sim \mc{N} \left(\mbf{0}, \mbf{Q}^u_k \right)$ and $\Exp \left(\cdot \right)$ is the exponential map for $SE(3)$. Figure~\ref{fig:process_snippet} shows how the \texttt{ProcessModel} interface can be utilized to define the process model, as well as the process model Jacobian and covariance.
The robot is also equipped with a UWB tag that collects range measurements to multiple anchors with known position. Denoting the $i^\mathrm{th}$ known anchor position in the world frame as $\mbf{a}_w^{i}$, the range measurements to the $i^\mathrm{th}$ anchor are in the form $y_k^i = g^i\left(\mbf{T}_k \right) + v_k^i$, where $v_k^i \sim \mc{N} \left(0, R_k^i \right)$, and the measurement model is given by 
    \begin{align} \label{eq:range_pose_to_anchor}
        g^i\left(\mbf{T} \right) = \norm{\mbf{r}_w - \mbf{a}_w^{i}}.
    \end{align}
Figure~\ref{fig:meas_snippet} shows how the \texttt{MeasurementModel} interface can be used to implement the range measurement model~\eqref{eq:range_pose_to_anchor}.

\begin{figure}[t]
    \centering
    \includegraphics[width=0.98\linewidth,clip,trim={0.5cm 0.7cm 0 0}]{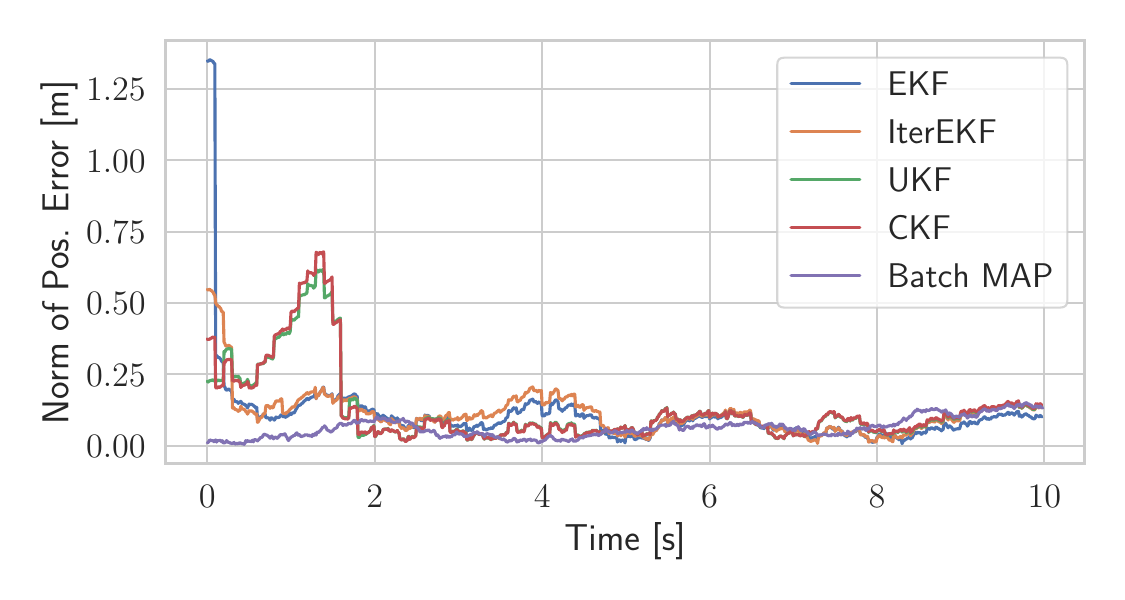}
    \vspace{-4mm}
    \caption{Norm of positioning error for ranging example, showcasing the ease of running various algorithms on arbitrary problems in \pynav.}
    \label{fig:single_range_rmse}
\end{figure}

\begin{figure}[t]
    \centering
    \includegraphics[width=0.98\linewidth,clip,trim={0.5cm 0.7cm 0 0}]{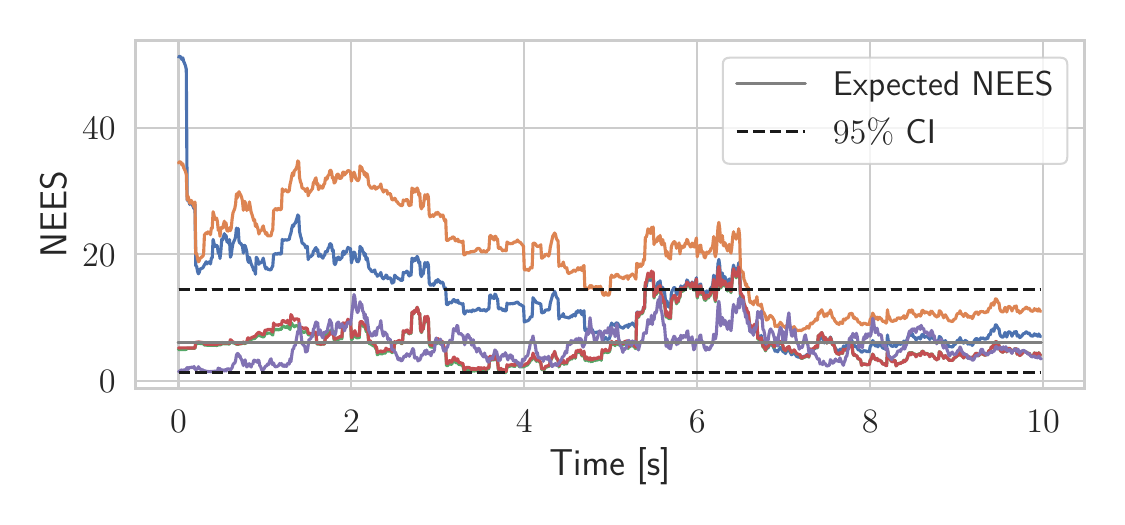}
    \vspace{-4mm}
    \caption{Normalized estimation error squared (NEES) for the sample problem, with the line colors corresponding to Figure \ref{fig:single_range_rmse}. \pynav provides utility functions for standardized computation and plotting of the NEES.}
    \label{fig:single_range_nees}
\end{figure}

Once the state definition, process model, and measurement models have been implemented for a particular problem, the user can then fuse inputs and measurements using any of the generic implementations of the algorithms outlined in Section~\ref{sec:on_manifold_estimation}. Figure~\ref{fig:single_range_rmse} shows the norm of the position error and Figure~\ref{fig:single_range_nees} shows the NEES for the considered sample localization problem with three anchors, for several different estimators. If groundtruth data is available, a collection of utilities included in \pynav can be utilized for error and consistency evaluation.  Note that, despite not being showcased here, the implementations of the estimation algorithms make it simple to handle measurements from varying sensors, at varying rates, as they all share a common interface.

\begin{figure}
    \centering
    \includegraphics[width=0.98\linewidth]{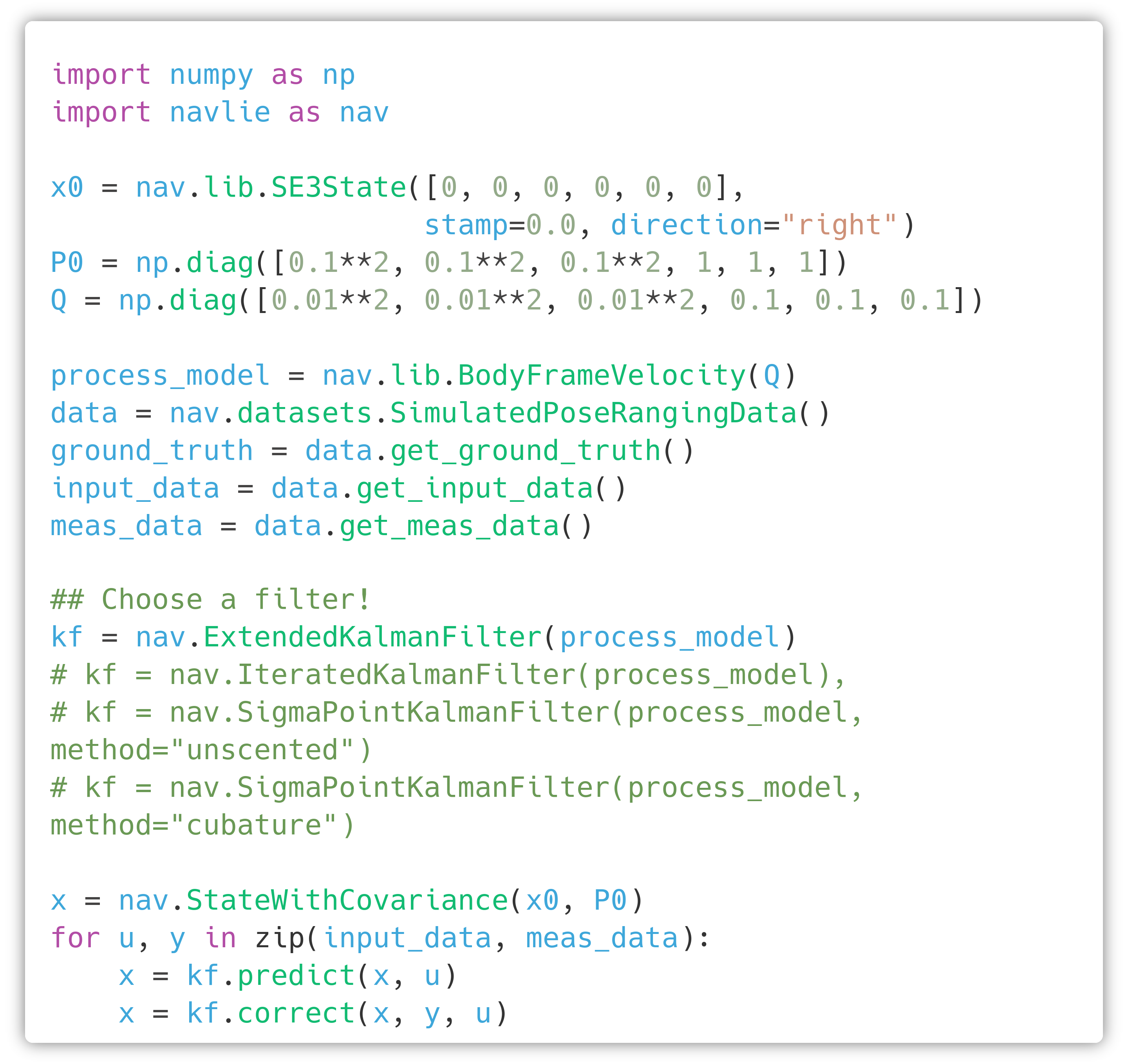}
    \vspace{-4mm}
    \caption{A typical top-level file using \pynav. Here, an EKF is chosen, but choosing a different algorithm is a matter of changing a single line of code.}
    \label{fig:filter_snippet}
\end{figure}


\section{Conclusion}
\label{sec:conclusion}
This paper presents \pynav, a Python package that provides a collection of on-manifold estimation algorithms and utilities suitable for a variety of state estimation  problems. \pynav allows for users to rapidly prototype algorithms, compare various estimation approaches, and analyze fundamental properties of the developed algorithms, such as estimator consistency. 



\vspace{-2mm}
\printbibliography


\end{document}